\documentclass{article}

\usepackage[english]{babel}
\usepackage{wrapfig}
\usepackage[letterpaper,top=2cm,bottom=2cm,left=3cm,right=3cm,marginparwidth=1.75cm]{geometry}

\usepackage{amsmath,amssymb}

\DeclareMathOperator{\E}{\mathbb{E}}
\DeclareMathOperator*{\argmax}{arg\,max}

\usepackage{graphicx}
\usepackage[colorlinks=true, allcolors=blue]{hyperref}
\graphicspath{{figures/}}

\title{Learning Visual Tracking and Reaching with Deep Reinforcement Learning on a UR10e Robotic Arm}

\author{
  Colin Bellinger\\
  Digital Technologies, National Research Council of Canada\\
  \texttt{colin.bellinger@nrc-cnrc.gc.ca}
  \and
  Laurence Lamarche-Cliche \\
  Software Engineering, Carleton University \\
  \texttt{LaurenceLamarcheClic@cmail.carleton.ca}
}

\date{}

\begin{document}
\setlength{\parskip}{10pt} 

\maketitle

\begin{abstract}
As technology progresses, industrial and scientific robots are increasingly being used in diverse settings. In many cases, however, programming the robot to perform such tasks is technically complex and costly. To maximize the utility of robots in industrial and scientific settings, they require the ability to quickly shift from one task to another. Reinforcement learning algorithms provide the potential to enable robots to learn optimal solutions to complete new tasks without directly reprogramming them. The current state-of-the-art in reinforcement learning, however, generally relies on fast simulations and parallelization to achieve optimal performance. These are often not possible in robotics applications. Thus, a significant amount of research is required to facilitate the efficient and safe, training and deployment of industrial and scientific reinforcement learning robots. This technical report outlines our initial research into the application of deep reinforcement learning on an industrial UR10e robot. The report describes the reinforcement learning environments created to facilitate policy learning with the UR10e, a robotic arm from Universal Robots, and presents our initial results in training deep Q-learning and proximal policy optimization agents on the developed reinforcement learning environments. Our results show that proximal policy optimization learns a better, more stable policy with less data than deep Q-learning. The corresponding code for this work is available at \url{https://github.com/cbellinger27/bendRL_reacher_tracker}

\end{abstract}

\section{Introduction}

Motivated by increasing labour costs and potential improvements in efficiency, a wide variety of scientific laboratories and  industries have adopted the use of robotic arms for production, maintenance and service \cite{ballestar2021impact}. Robots are often ideal for repetitive, strenuous, and complex tasks that require speed and precision. To achieve this, however, the current generation of industrial and scientific I robots require manual programming by robotics engineers. The programmer uses motion planning based on forward and inverse kinematics, where forward kinematics combine the direct measurement of all joint orientations with the lengths of linkages connected to the joints to determine the end-effector position and inverse kinematics establish the values of joint positions required to place the end-effector at a new location with a new orientation. To make the incorporation of industrial robots more inclusive, modern systems enable skilled operators to manual manipulation of the robot via a control pendant. Although this can simplify the process, online manipulation via the control pendant presents its own challenges and limitations \cite{lobbezoo2021reinforcement}. Small and medium-sized institutions face additional barriers to the incorporation of robotic arms related to the cost of implementation and reprogramming the robot to handle evolving requirements \cite{pedersen2016robot}. 

Reinforcement learning (RL) is a form of machine learning, which is applicable to sequential decision-making problems, such as navigating \cite{chiang2019learning}, control \cite{Yufeng2022}, complex games \cite{silver2018general}, \textit{etc}. With RL, the agent learns a control policy to complete a task or achieve some goal. Importantly, the learning setup does not require supervised learning and the task can evolve or be adapted as requirements and environments change. As a result, RL offers a general alternative to task-specific robot programming with the potential to simplify the adoption of industrial robots. On the other hand, many challenges exist in the applications of RL to real-world robotics. These include the requirement to ensure the safety of the robot and surroundings during learning, the need to define a reward function to effectively direct policy learning, and potentially long training times \cite{sunderhauf2018limits}. 

\begin{wrapfigure}{r}{0.5\textwidth}
    \centering
    \includegraphics[width=0.45\columnwidth]{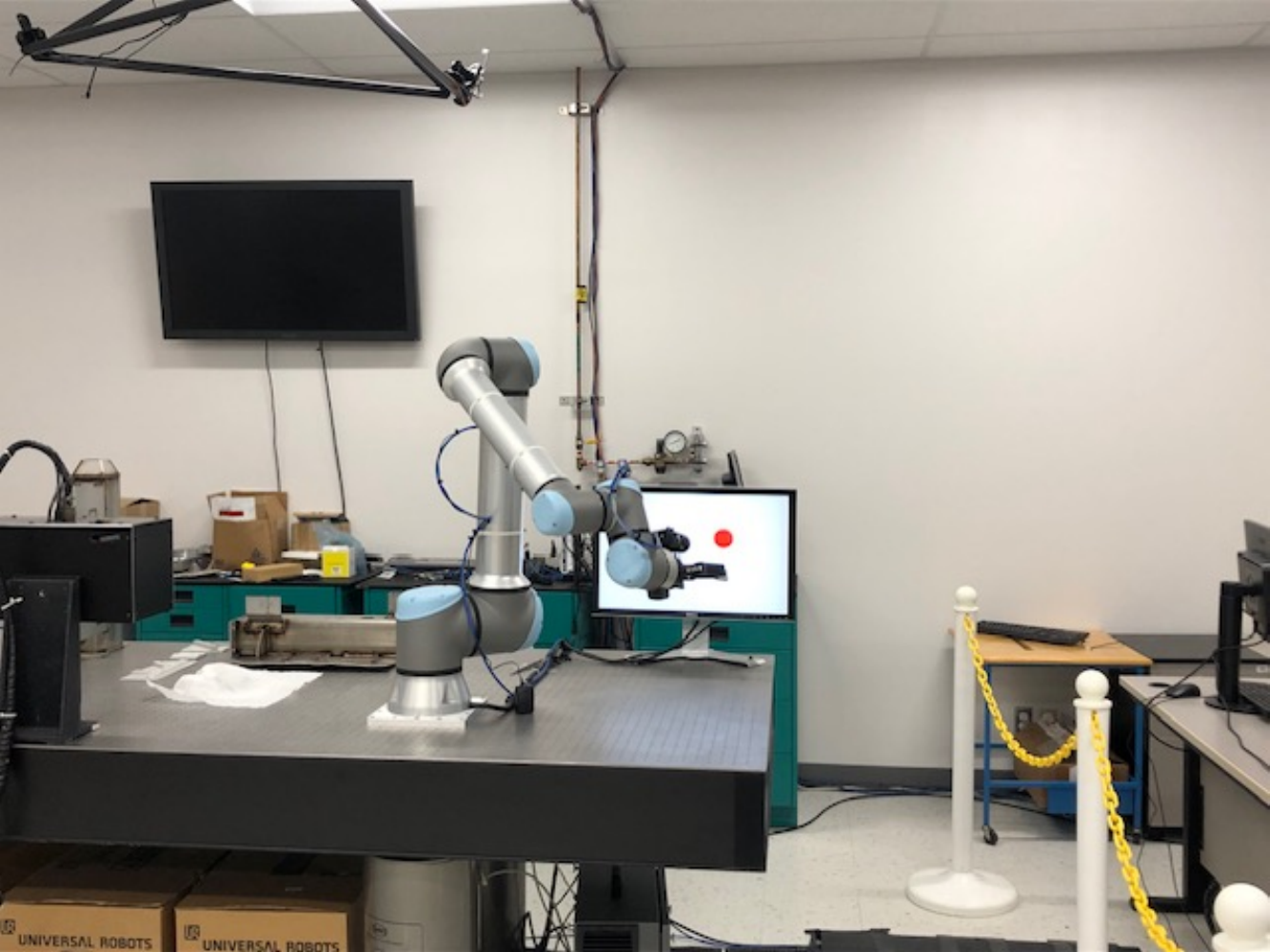}
    \caption{UR10e Robot used in this work.}
    \label{fig:UR10E}
\end{wrapfigure}

In this work, we evaluate the ability to learn image-based reaching and tracking asks via reinforcement learning on the UR10e robotic arm by Universal Robots. The reaching and tracking tasks considered here are inspired by previous work in \cite{Yufeng2022} with a UR5 robot. The UR10e robot used in this work is shown in Figure \ref{fig:UR10E}. In our experiments, we evaluate the potential of the off-policy, value-based method Deep Q-Learning (DQN), and the on-policy, policy gradient method Proximal Policy Optimization (PPO). Our results show that PPO learns a better, more stable policy with less data than DQN. In order to improve its performance, DQN requires a more sophisticated exploration strategy and non-uniform sampling of its replay buffer to focus updates on the most informative experience. 

\section{Background}

RL problems are formally described by a Markov decision process (MDP). An MDP is defined by the tuple $(S,A,P,R,\gamma)$, where $S$ and $A$ are discrete or continuous  action- and state-spaces, $P$ is the state transition dynamic which map state-action pairs at time $t$ onto a distribution over next states at time $t+1$, $R$ is a reward function that emits an instantaneous reward at time $t+1$ given the state-action pair at time $t$, and $\gamma \in [0,1)$ is a discount factor that balances the importance of immediate versus future rewards. Setting $\gamma$ closer to 0 place greater importance on rewards accumulated in the near-term. In RL, $P$ and $R$ are unknown to the agent. The states under the MDP setup satisfy the Markov Property, which dictates that the future is conditionally independent of the past given the present state. If this property does not hold, then the learner faces a partially observable Markov decision process (POMDP). In addition to the elements of the MDP, A POMDP contains an observation space $\Omega$, which alone is not sufficiently informative to select the next action, and a state-action observation probability function $O(o_{t+1}| a_t, s_{t+1})$. When RL is applied to POMDPs, $\Omega$ and $O$  are also unknown to the agent. The agent receives an observation at each time step, $t$, rather than the underlying state of the system. Image-based RL applications are often partially observable because a single image is not sufficient to encode the dynamics of the world and other details that are relevant to action selection. In deep RL, POMDPs are typically handle with a combination of stacking a sufficiently long history of state-action pairs, a state encoder and/or a recurrent neural network \cite{igl2018deep}. 

In RL, the agent learns a policy, $\pi$, through sequential interactions with the environment (the world outside of the agent). At each time step, the policy maps the state to an action or distribution over actions $\pi: s_t \rightarrow a_t$. The sequence of states, actions and rewards, $(s_0, a_0, r_1, s_1, a_1, r_2, s_2,...)$ under a policy $\pi$ is denoted a trajectory or roll-out. The agent's objective is to learn a policy $\pi^*$ that maximizes the return, $G$, 
\begin{equation}
    \pi^* = \argmax_\pi \E[G|\pi].
\end{equation}
The return, $G$, is defined as: 
\begin{equation}
    G_t = \sum_{t=0}^{t=T-1} = \gamma^t r_{t+1},
\end{equation}
where $T$ is the length of a finite horizon MDP, or $\infty$ for an infinite horizon MDPs. In the case of infinite horizons $\gamma\in[0,1)$. For more details see \cite{sutton2018reinforcement}. 

\section{Related Work}

\subsection{Reinforcement Learning}

In recent years, the combination of reinforcement learning with deep feature learning has produced impressive advancements in the field of RL. Some noteworthy successes in deep RL include agents learning to play Atari 2600 video games directly from images \cite{mnih2015human}, learning to play and defeat the world champion of Go \cite{silver2016mastering}, and learning to cool large industrial buildings \cite{luo2022controlling}.

Deep Q-Learning (DQN) provided an early breakthrough in deep reinforcement learning \cite{mnih2013playing}. It was the first deep RL algorithm capable of effectively learning directly from raw visual inputs. DQN is a value-based, off-policy RL algorithm that is suitable for discrete action spaces and discrete or continuous state spaces. DQN utilizes a deep neural network as its state-action value function ($Q$-function). More specifically, given a state $s_t$, the neural network $Q$-function predicts the value of taking each possible action $a\in A$ in $s_t$, where the value approximates the expected return starting from $s_t$, taking the action $a$, and thereafter following policy $\pi$. During training, the agent interacts with the environment according to an $\epsilon$-greedy policy. Specifically, $1 - \epsilon$ percent of the time, the agent selects an action according to its current policy, and $\epsilon$ percent of the time the agent selects a random action. The aim is to gain experience whilst balancing the exploration-exploitation trade-off. Tuples of experience are recorded in a fixed-length experience replay buffer and used to update the $Q$-function according to the objective: 
\begin{equation}
    L_i(\theta_i) = \E_{(s,a,r,s^\prime)} \bigg[ \bigg(y_i - Q(s,a;\theta_i)\bigg)^2\bigg] 
\end{equation}
with
\begin{equation}
    y_i = r + \gamma \max_{a^\prime} Q(s^\prime, a^\prime;\theta^-),
\end{equation}
where $\theta$ is the $Q$ network parameters,  $\theta^-$ is the target network parameters and $i$ is the current iteration of network updates. The parameters of the target network are copied from the $Q$ network every $\tau$ step or slowly shifted towards $\theta$ using soft Polyak updating. This, along with the use of the replay buffer, improves the stability of the deep $Q$ network. Other improvements to DQN include Double $Q$-Learning, which improves DQN's overestimation action values under certain conditions \cite{van2016deep} and Dueling DQN which generalizes learning across actions leading to better policy evaluation in the presence of many similar-valued actions \cite{wang2016dueling}, and prioritize experience replay which improves learning by replaying important transitions more frequently \cite{schaul2015prioritized}. A key advantage of off-policy methods, such as DQN, is sample efficiency. In comparison to on-policy methods, off-policy methods gain sample efficiency through the reuse of data collected under previous versions of the policy and the option to curate training batches such that experiences that produced more reward in the past are seen more frequently during training. 

Proximal Policy Optimisation (PPO) is a policy gradient form of RL algorithm that has many of the advantages of its predecessor, Trust Region Policy Optimization (TRPO), but is easier to implement, more general, and has better sample efficiency \cite{schulman2017proximal}. Unlike DQN, PPO is suitable for discrete or continuous state- and action- spaces. A key feature of PPO is the use of clipping in the objective function to reduce the risk of large policy updates that lead to a collapse in performance. Policy updates in PPO are conducted in an on-policy manner by collecting experience roll-outs under the current parameterized policy $\pi_{\theta_i}$ and updating the policy as:
\begin{equation}
    \theta_{i+1} = \argmax_{\theta} \mathbb{E}_{s,a \sim \pi_{\theta_i}} \big[ L(s,a,\theta_i,\theta)\big]
\end{equation}
using stochastic gradient ascent. The loss, $L$ is:
\begin{equation}
     L(s,a,\theta_i,\theta) = \min\bigg( \frac{\pi_\theta(a|s)}{\pi_{\theta_i(a|a)}} A^\pi_{\theta_i}(s,a), \text{clip} \big(\frac{\pi_\theta(a|s)}{\pi_{\theta_i(a|a)}}, 1-\epsilon,1+\epsilon \big)A^\pi_{\theta_i}(s,a) \bigg),
\end{equation}
where $A$ is an estimate of the advantage function, which measures whether or not an action is better than the policy's default behaviour, and  $\epsilon$ is a small hyper-parameter that controls the maximum step size of the policy update. A key advantage of PPO over many other RL algorithms is its robustness under a wide range of hyper-parameters. Particularly in comparison with DQN, PPO can require significantly less hyper-parameter tuning. This is a beneficial feature for many RL applications, such as robotics. 

Soft Actor Critic (SAC) is an off-policy algorithm that is suitable for continuous state- and action-spaces \cite{haarnoja2018soft}. In addition, there is a variation applicable to discrete action-spaces \cite{christodoulou2019soft}. SAC learns a stochastic policy that is trained to maximize the expected return with entropy regularization. In particular, it learns a parameterized policy $\pi_\theta$ and Q-networks $Q_{\phi_1}$ and $Q_{\phi_2}$. SAC uses the clipped double-Q trick and takes the minimum Q-value of $Q_{\phi_1}$ and $Q_{\phi_2}$ for use in the loss function:
\begin{equation}
    L(\phi_i, D) = E_{(s,a,r,s^\prime,d)\sim D} \bigg[\bigg(Q_{\phi_i}(s,a) - y(r,s^\prime,d)\bigg)^2  \bigg],
\end{equation}
where $D$ is the replay buffer, $d$ indicates if $s^\prime$ is terminal, and:
\begin{equation}
    y(r,s^\prime,d) = r + \gamma(1-d)\bigg( \min Q_{\phi_{target,j}}(s^\prime, \Tilde{a}^\prime) - \alpha\text{log}\pi_\theta(\Tilde{a}^\prime|s^\prime)\bigg), ~~~ \Tilde{a}^\prime \sim \pi_\theta(\cdot | s^\prime),
\end{equation}
where $\alpha$ controls the explore-exploit trade-off. The $\alpha$ value is the main hyperparameter that should be tuned in SAC. A larger $\alpha$ encourages more exploration. The policy network, $\pi_\theta$, acts to maximize the expected future return plus expected future entropy. In this way, both future rewards and exploration are encouraged. In particular, SAC learns to maximize:
\begin{equation}
    V^\pi(s) = \E_{a \sim \pi}\big[Q^\pi(s,a)\big] + \alpha  H(\pi(\cdot | s)),
\end{equation}
where $H(\cdot)$ is entropy. In practice, this is optimized using the reparameterization trick where a sample is drawn from $\pi_{\theta}(\cdot|s)$ by computing a deterministic function of state, policy parameters, and independent noise. In addition to the sample efficiency that comes with off-policy learning, SAC has the advantage of being relatively insensitive to hyperparameters and balances the exploration-exploitation trade-off through the use of entropy regularization. 

\subsection{Reinforcement Learning with Physical Robots}

Advancements in RL and the reduced costs of some robotics systems have resulted in a significant increase in the number of RL researchers working on robotics applications. This includes, for example, object manipulations skills, \cite{gu2017deep,haarnoja2018soft}, peg insertion  \cite{lee2019making}, targeted throwing \cite{ghadirzadeh2017deep}, and dexterous manipulation \cite{zhu2019dexterous}. Nonetheless, many significant challenges remain for the widespread application of RL to robotics. These include issues such as image observation and multi-format observation, sample efficiency, sim-to-real training, human-free learning, including safe and reset-free learning, generalization across tasks, and learning in the open world. 

In relation to this work, there have been a few other applications of RL to versions of Universal Robots' UR cobotic arms. This work takes direct inspiration from Yufeng and Mahmood \cite{mahmood2018setting,Yufeng2022} work on asynchronous RL for real-time control. Yufeng and Mahmood trained SAC RL to control a UR5 robot to complete reaching and tracking tasks. In their experimental setup, the target is rendered on a computer monitor in front of the robot. Rendering the target on a computer monitor has many practical advantages, such as safe and reset-free learning, whilst maintaining some important real-world robotics challenges. The state-space in this setup is composed of visual inputs from a camera mounted on the end-effector and the joint position of the arm. The combination of images and joint information reduces the risk of partial observability. In \cite{meyes2017motion}, RL is applied to learn a control policy to enable a UR5 robot to complete the wire rope problem. The authors in \cite{abs-2005-02632}, compared Trust Region Policy Optimization and Deep Q-Network with Normalized Advantage Functions to Deep Deterministic Policy Gradient and Vanilla Policy Gradient in simulation and with transfer to UR5. UR3e robots were the subject of RL learning the peg in the hole task in \cite{beltran2020variable}, pick and place in \cite{gomes2022reinforcement} and goal-based RL \cite{parak2021comparison}.

\section{RL Environment}

In the following experiments, a UR10e robot from Universal Robots learns to carry out reaching and tracking tasks from visual inputs. The inputs are captured by a Point Grey camera mounted on the end-effector of the robotic arm. Tracking and reaching are performed with respect to a digital target rendered on a 1920x1200 monitor fixed at a distance of 70 cm from the robot. The target is a red circle on a white background. The setup is shown in Figure 1. In order to obtain results that would transfer more easily to real-world applications, the background of the room is unobstructed and total light in the room can vary due to changes in the outdoor light through the North facing windows. The specifics of the physical reacher and tracker RL environment are described in the following sections. 

In both the reacher and tracker tasks, the robotic arm has a fixed start position at the beginning of each episode. In the reacher task, the target location can be set to fixed or be randomly reset at the beginning of each episode. Alternatively, in the tracker task, the target is initialized at the centre of the monitor at the beginning of each episode. The target slowly drifts in one of 4 coordinate directions out from the centre of the monitor at a rate of 3 pixels per RL time step. The drift direction of the target is uniformly sampled at the beginning of each episode. The monitor frame boundaries reflect the target by reversing the direction on one axis.

\subsubsection{Action Space}

The UR10e robotic arm has 6 degrees of freedom (base, shoulder, elbow and 3 wrist joints). The action space is composed of 10 discrete actions. These rotate one of of 5 joints either clockwise or counterclockwise (the third wrist would only rotate the camera, and therefore it was not included in the action space). Each joint's rotation actions are set proportionally to the size of the joint, with larger joints having slightly larger rotation actions. In particular, the base rotates 0.025 radians (rads), the shoulder and elbow rotate 0.02 rads, and wrists 1 and 2 rotate 0.015 rads. The agent selects one joint to move per time step in either the positive or negative direction from its current position. 

For safety purposes, prior to sending a selected action to the robot via the UR10e's movej API command, the RL environment validates the outcome of the move. The validation checks if the move will cause the arm to move out of the defined Cartesian limits. If the action is deemed to violate the limits, the environment steps forward without the robot moving. 

\begin{figure}[tb]
\centering
\includegraphics[width=0.8\columnwidth]{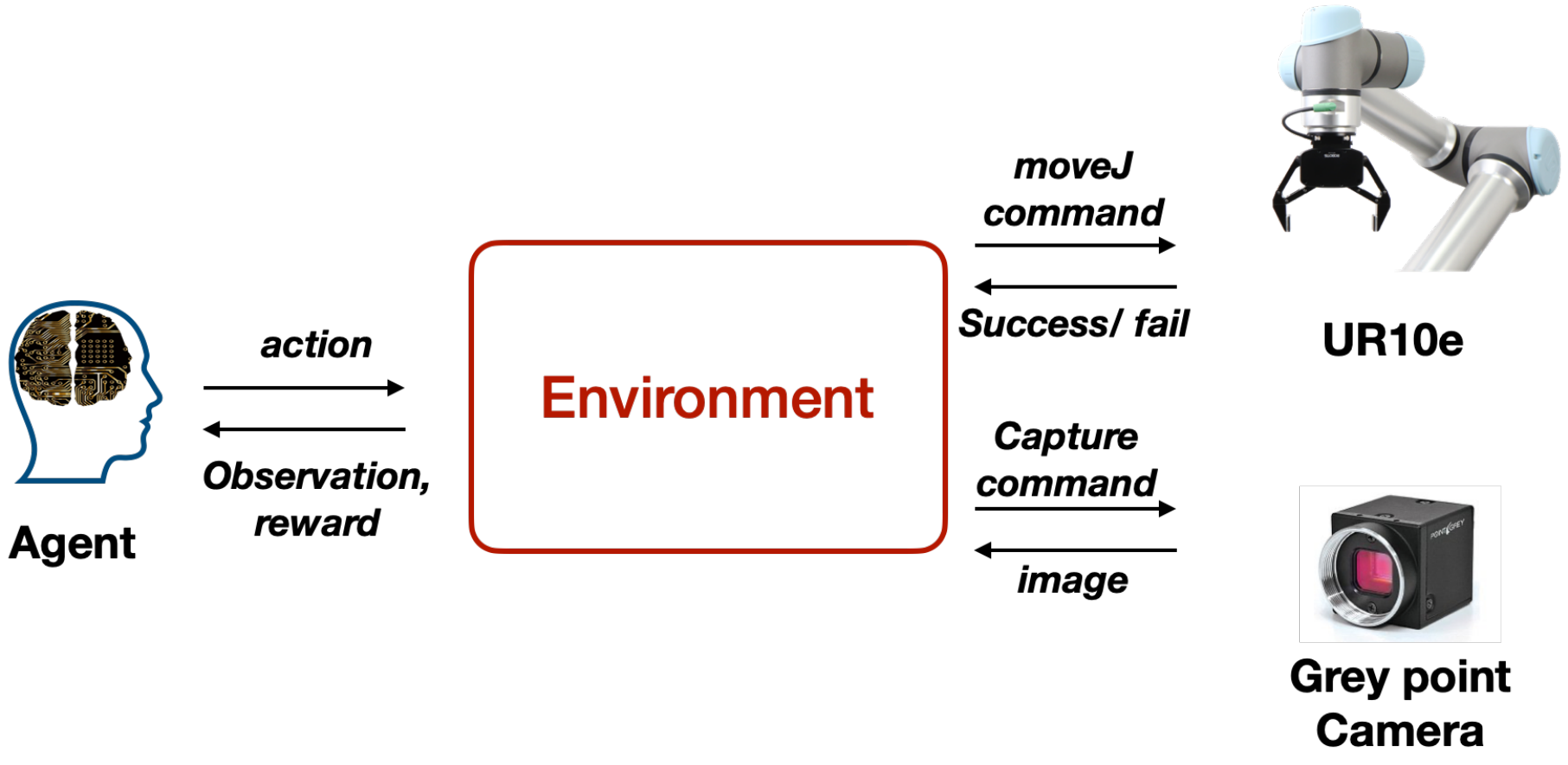}
\caption{This figure displays the agent-environment-hardware interaction framework. }
\label{fig:framework}
\end{figure}

\subsubsection{Dynamics of the Environment}

The environment is episodic. One time step in the environment includes one synchronous interaction between the agent and the environment. In particular, this involves: 1) the environment calculating the current observation and reward, 2) sending the observation and reward to the agent, 3) the agent selecting the next action, 4) the agent sending the action to the environment, and 5) the environment commanding the appropriate move of the robot. The interaction framework is visually presented in Figure \ref{fig:framework}.

Each episode starts the robotic arm in a fixed joint position within the Cartesian boundaries. The episode ends when the agent has achieved the goal. For training purposes, we truncated the episode after 150 time steps if the agent as not achieved the goal. 

If the agent selects an action that will trigger a protective stop due to a rotation or boundary limit, the action is not initiated. Thus, the agent remains in its current position and state does not change. An observation from this location is returned to the agent. All other actions cause the arm to move a predefined amount via the movej command. The camera observation is made after the move is completed.

\subsubsection{Reward}

The goal in each task is for the agent to learn to move camera on the end of the robotic arm as close as possible to the target and centre the target in the camera frame. A shaped rewards schemes is evaluated for this purpose. Arrival at the goal is determined by the size and centering of the target in the current observation frame. Given the frame size of 400x300 and that the target is included in the observation, the maximum distance between the centre of the frame and the centre of the visible target is 250 pixels. The maximum radius of the target is 40 pixels and the minimum radius is 10 pixels. For the goal to be achieved, the target must have a radius greater than 30 pixels and be less than 70 pixels from the centre of the frame. 

In the shaped reward setting, the agent receives a reward of 20 for achieving the goal. A reward of -0.01 is given if the target is not in the frame. If the target is in the frame, but the goal has not been achieved, intermediate rewards between zero and one are given. The intermediate rewards are increased based on how centred the target is and how close the camera is to the target. These are calculated as:
\begin{equation}
    \text{dist\_reward} = -1 \times \text{dist\_to\_goal} / \text{max\_dist\_to\_goal}
\end{equation}
\begin{equation}
    \text{radius\_reward} = (\text{target\_radius} -  \text{max\_radius})/(\text{max\_radius}-\text{min\_radius})
\end{equation}
\begin{equation}
    \text{reward} = \text{radius\_reward} + \text{dist\_reward}
\end{equation}

If the agent selects an action that will trigger a protective stop due to a rotation or boundary limit, 1 is subtracted from the reward in order to discourage the agent from hitting the limits. 

\subsubsection{Image Processing}

\begin{figure}[tb]
\centering
\includegraphics[width=0.95\columnwidth]{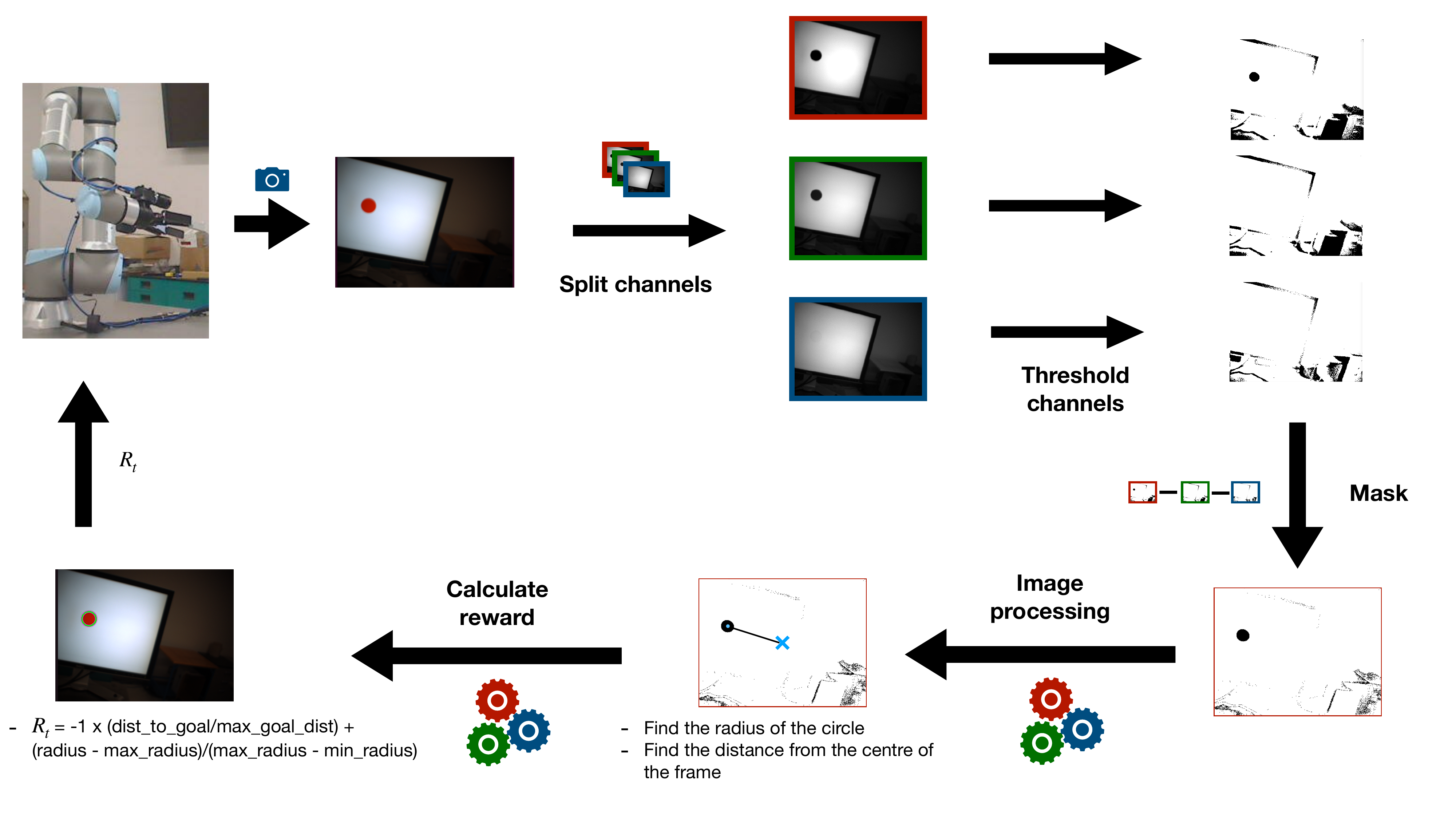}
\caption{This figure displays the image processing steps to determine if the target is in the current frame, along with the size and location of the target if it is present. }
\label{fig:image_proc}
\end{figure}

A significant challenge in RL with image observations is calculating the reward. In this setup, the reward is based on the size and centring of the target (red circle) in the image captured by the camera on the robotic arm. Therefore, in order to compute the reward at each time step, the environment uses image processing to determine if the target is in the current frame. If the target is in the frame, the environment determines the radius and location of the target to be used in the reward calculation discussed above. 

There are multiple approaches that could be used to detect and quantify the target circle. In this work, we use openCV's HoughCircles function, which uses the Hough transform \cite{duda1972use}, to identify circles in the observation image at each time step. The sensitivity and specificity of circle detection are controlled by four hyperparameters. In the initial experiments, the HoughCircles function was optimized on raw image observations. The experimental setup includes many factors that can be expected in open-world robotics, such as variable lighting, background noise containing a mix of objects, textures and colours, and dynamically changing angles and distances between the camera and target. Each of these factors contributes to poor sensitivity and specificity with HoughCircles function. In addition, the background has circular objects that needed to be ignored during the target identification for reward calculation. Therefore, the RL algorithm must be robust to some reward signal noise.

As a result, a multi-step process to simplify the image observation by removing background details is used. This serves to significantly reduce false positives in the reward calculation. The steps are presented in Figure \ref{fig:image_proc}. First, the observation image is split into its three component channels (red, green and blue). Each channel has a threshold applied which maps all pixels, except those with nearly total presence, to total absence. The result of this is that most background details are removed, and in the case of the blue and green channels, the target circle is also removed. This can be seen in the images after the thresholding step in the figure. Next, the thresheld blue and green channels are applied as masks to the thresheld red channel. The resulting red channel has nearly all of the background details removed, with the target clearly intact. The final steps are to apply the HoughCircles function and calculate the reward as described in the previous subsection. 

\section{Experimental Setup}

We assess the performance of PPO and DQN on the reacher and tracker environments\footnote{ The corresponding code for this work is available at \url{https://github.com/cbellinger27/bendRL_reacher_tracker}}. In each case, the agent is trained end-to-end from scratch. Each agent is evaluated over 3 independent trials. PPO is trained over 40k time steps and DQN is trained over 60k time steps using the same CNN architecture. DQN uses a replay buffer of 20k and linear decay in the $\epsilon$ value from 1 to 0.01.

The results are graphically presented as the mean and standard error of reward per episode step. In addition, the steps per episode over the course of learning is plotted as an alternative perspective on performance. Fewer steps per episode indicate better performance.

All of the experiments utilize a HP Z6 G4 Workstation including Intel Xeon(R) Silver 4214R CPU @ 2.40GHz $\times$ 48 with the Ubuntu 22.04 operating system. Universal robot UR10e is running the URSoftware version 5.12. Communication between the Linux desktop an the UR10e robot uses ur\_rtde 1.5.5 software\footnote{\url{https://pypi.org/project/ur-rtde/}}. 

\section{Results and Discussion}

\begin{figure}[tb]
\centering
\includegraphics[width=0.48\columnwidth]{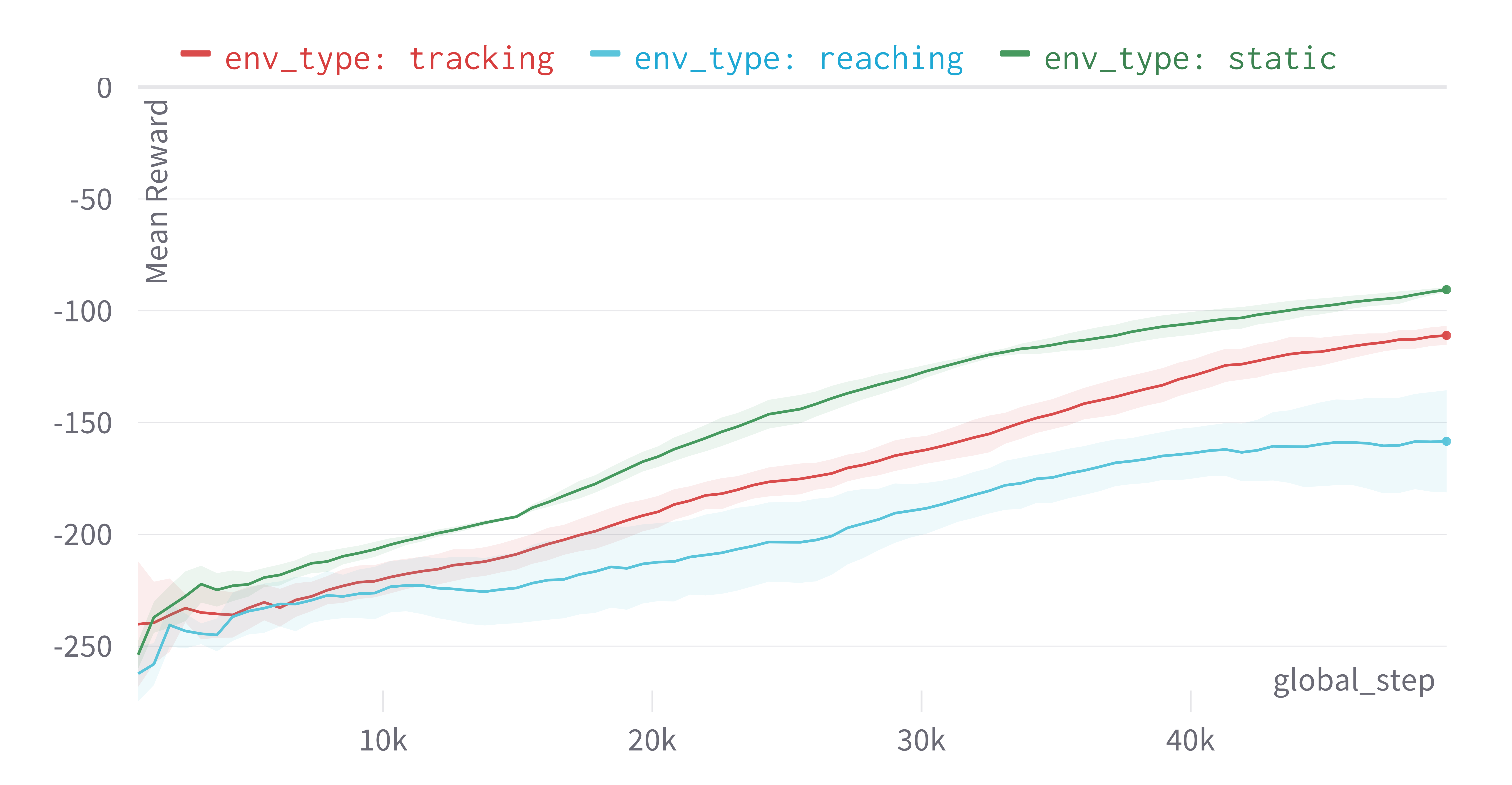}
\includegraphics[width=0.48\columnwidth]{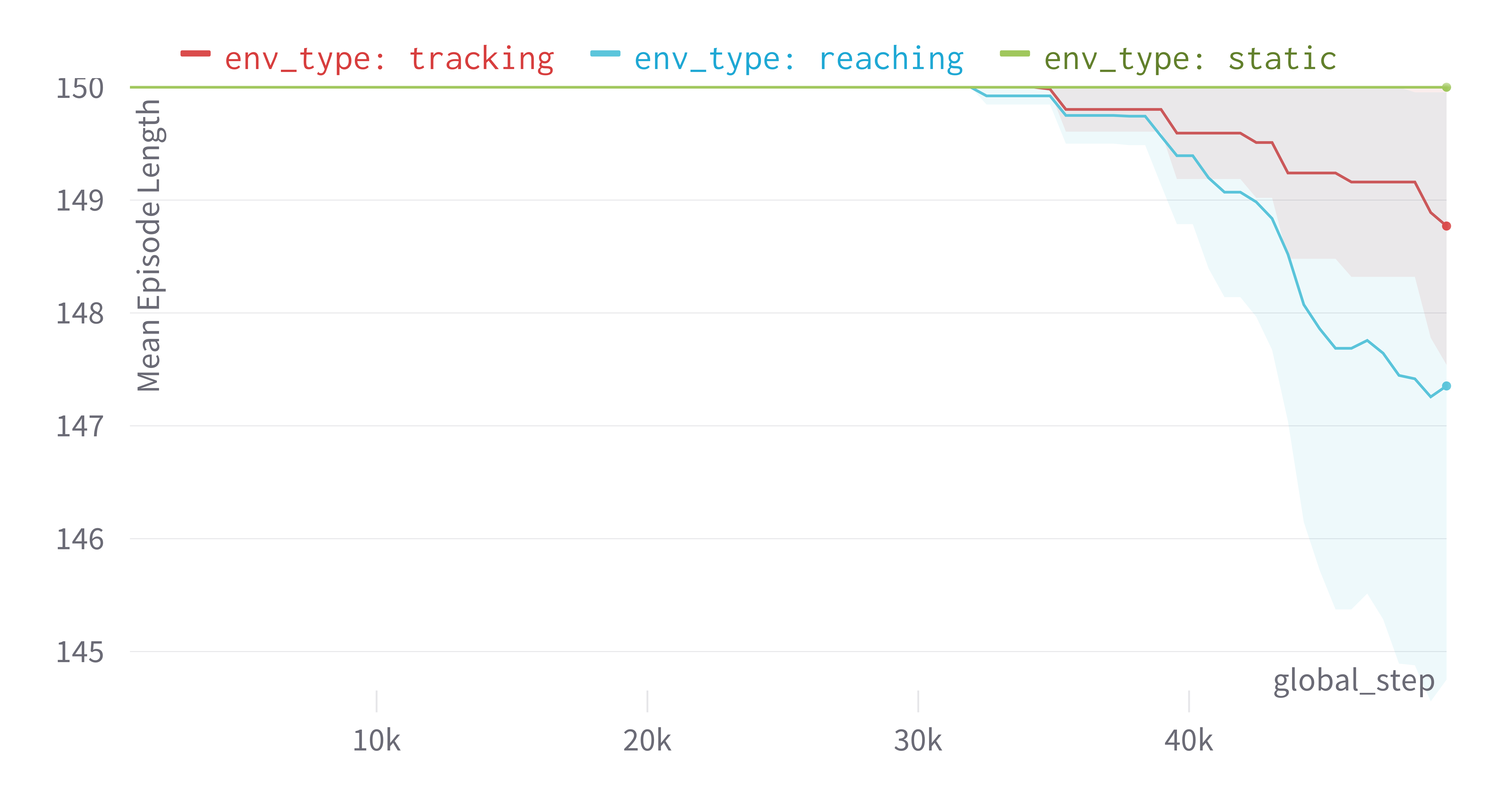}
\includegraphics[width=0.48\columnwidth]{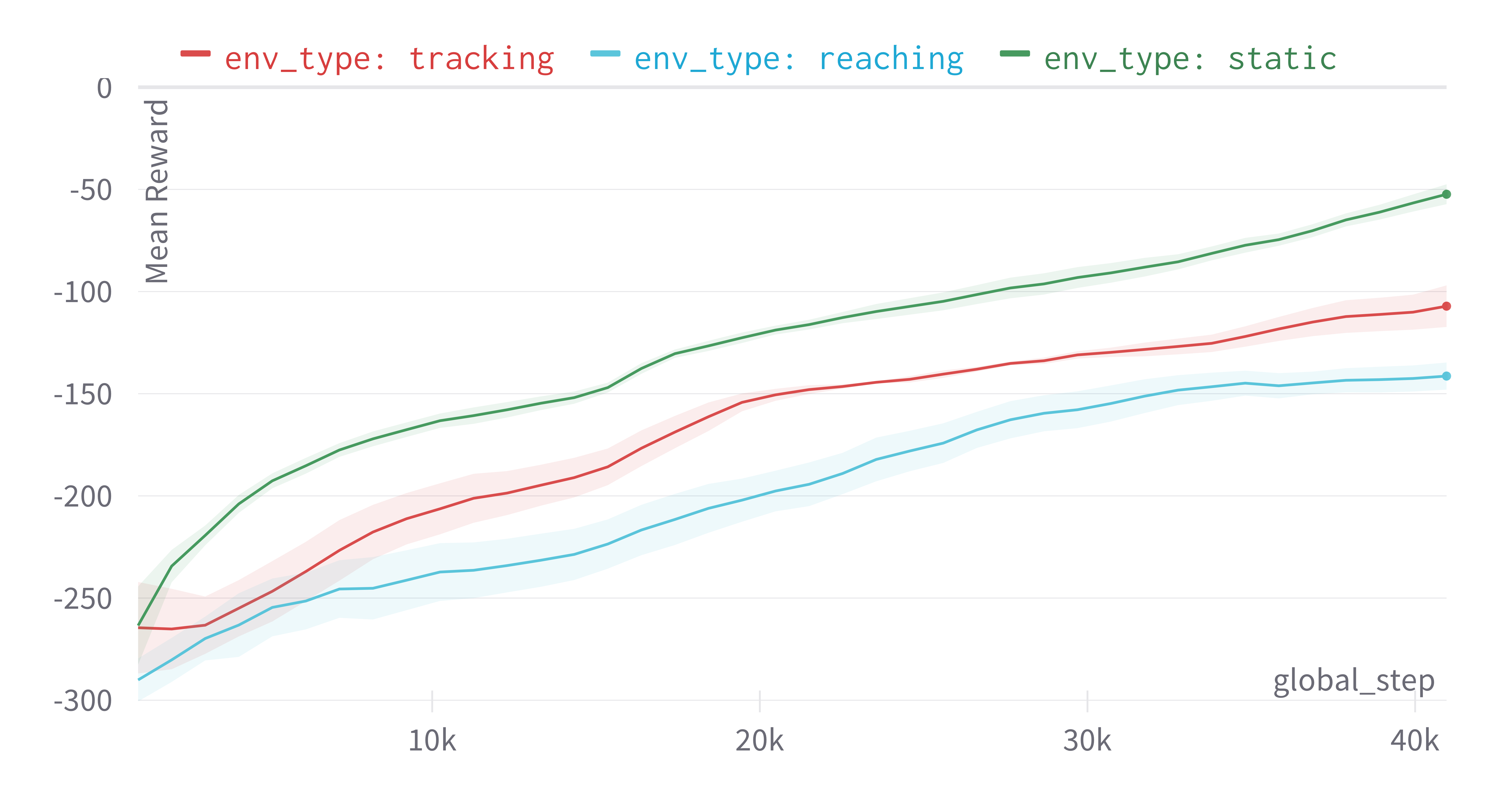}
\includegraphics[width=0.48\columnwidth]{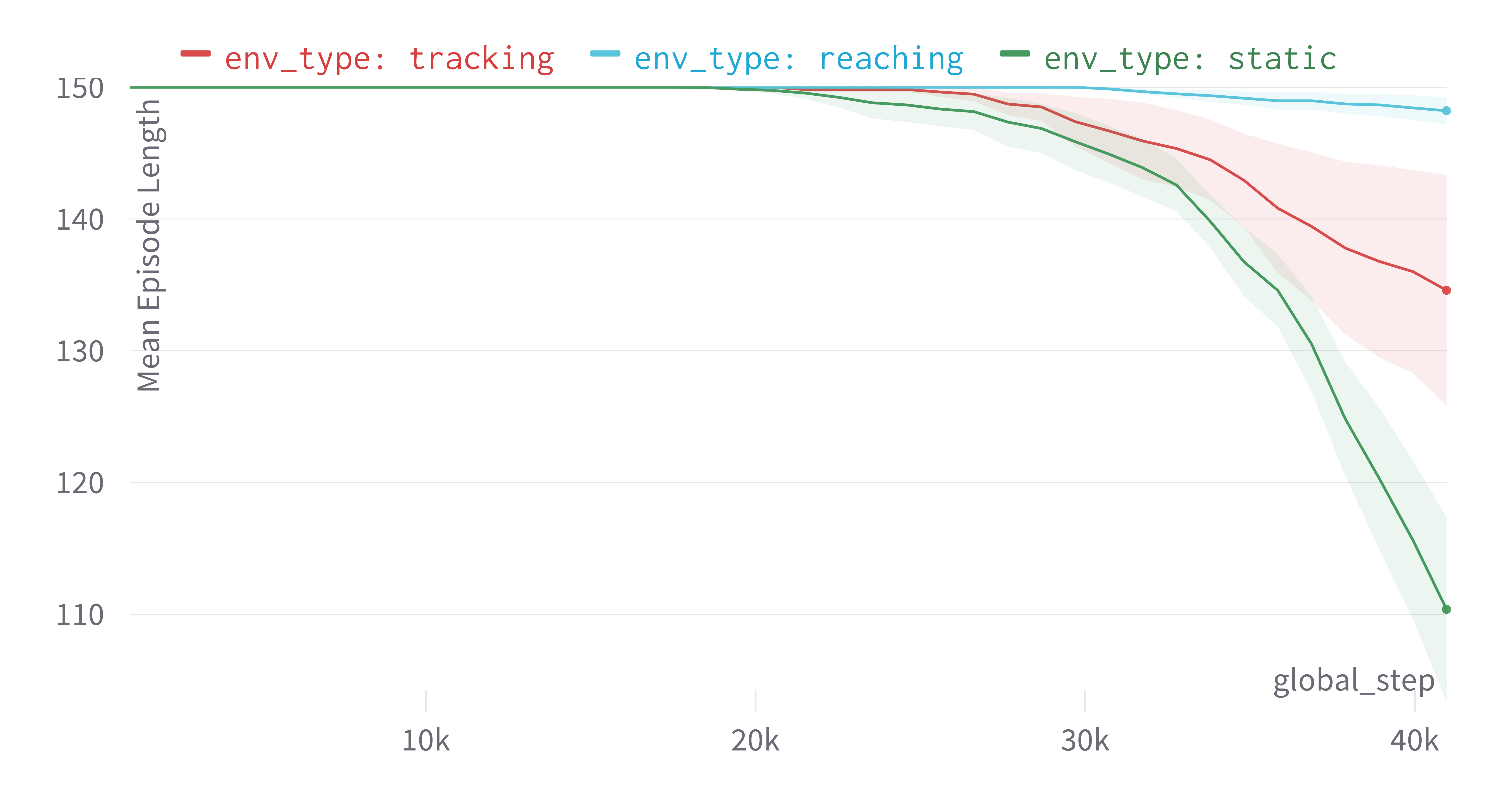}
\caption{DQN (top row) and PPO (bottom row) results. The left plots in each row depict the mean rewards on the tracking, reaching and static reaching environments. The right plots in each row present the number of steps per episode in each environment.}
\label{fig:results}
\end{figure}

Figure \ref{fig:results} includes the learning curves for DQN and PPO on each environment in terms of mean reward (left) and mean steps per episode (right). The plots show that both RL algorithms improve their policies over the course of training to increase the mean reward. After 40k steps, however, PPO has a much higher mean reward than DQN on all three tasks. After an additional 20k of training, DQN is approximately equivalent on reaching and tracking, and remains much worse than PPO (with 40k of training) on the static reaching task. This suggests that epsilon greedy exploration and uniform sampling from DQN's replay buffer are likely causing slow learning. Thus alternative strategies are needed to improve exploration and focus policy updates on the most valuable experience. 

The episode length plots demonstrate if, and how quickly, the agents are able to achieve the goal on each task. The most striking result is that although the reward results show that DQN learns to focus on the target in the static reacher task, it fails to achieve the goal within the maximum episode length of 150 steps. Alternatively, PPO learns to achieve the goal in a mean of 110 steps and is still improving its policy when training was stopped. Given that PPO has superior performance, the remainder of the analysis focuses on it.

In comparing the performance of PPO across the three environment types, the plots illustrate that the agents are best at the static reacher, where the target is always located at the centre of the monitor. Alternatively, the reacher environment, where the target is randomly re-positioned at the beginning of each episode, is the most challenging of the three environments. The tracker environment, which has the target start each episode at the centre of the monitor and slowly drifts away, falls between static reacher and reacher in terms of complexity for PPO. The ordering of performance is likely due to the fact that in static reacher and tracker, the target is always initialized in the same location. As a result, the agent can quickly learn the general area that it should move towards at the beginning of each episode. Alternatively, because the target is randomly positioned at the beginning of each reacher episode, the agent must search for and find the target before it can move it on the target. Due to the nature of on-policy learning, the agent may be biased to search areas where the target occurred in the more recent updates and slowly forget pasted experience. In terms of training time, the clock time is approximately 3:30 hours for each agent.

\begin{figure}[tb]
\centering
\includegraphics[width=1\columnwidth]{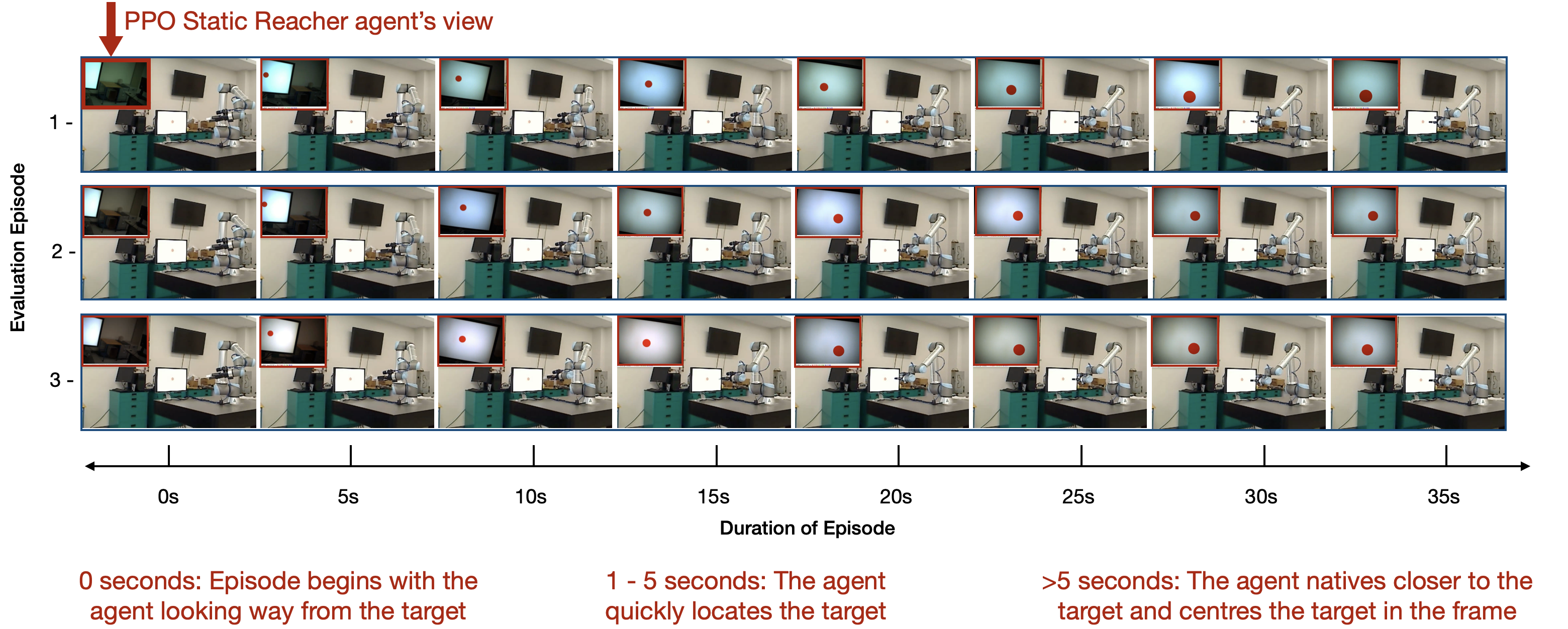}
\caption{This figure displays three policy roll-outs for PPO on the static reacher environment after training is completed. It demonstrates that the agent learned to find the target and then navigate toward the target while keeping the target centered in the field of view. }
\label{fig:ppo_sreacher_frames}
\end{figure}

\begin{figure}[tb]
\centering
\includegraphics[width=1\columnwidth]{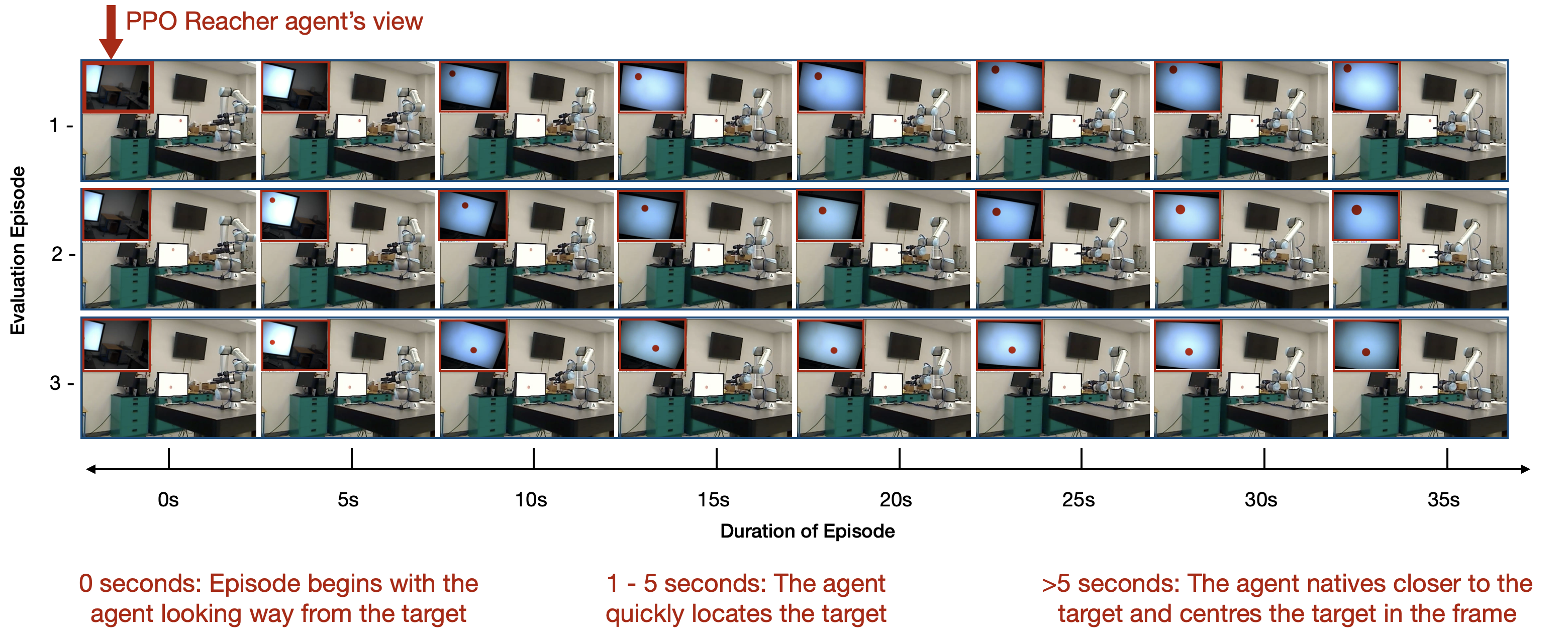}
\caption{This figure displays three policy roll-outs for PPO on the reacher environment after training is completed. It demonstrates that the agent learned to find the target and then navigate toward the target while keeping the target centered in the field of view. }
\label{fig:ppo_reacher_frames}
\end{figure}

\begin{figure}[tb]
\centering
\includegraphics[width=1\columnwidth]{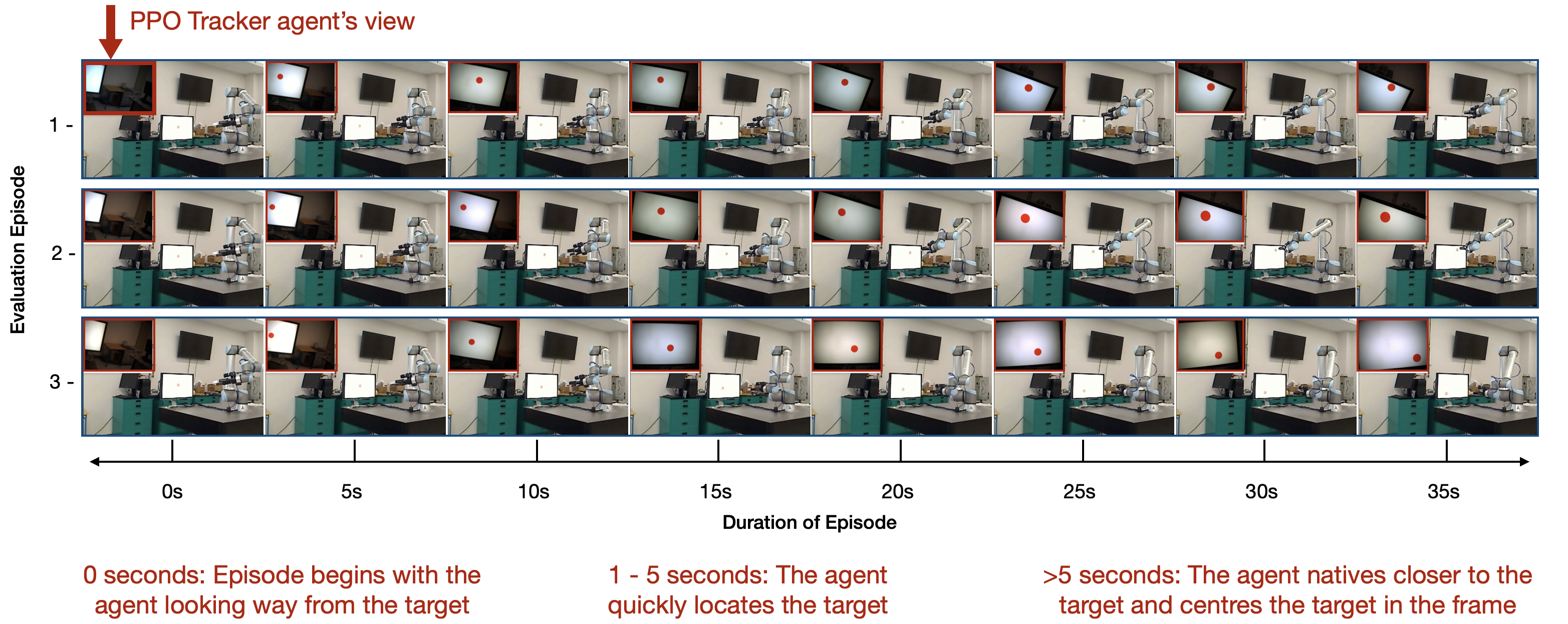}
\caption{This figure displays three policy roll-outs for PPO on the tracker environment after training is completed. It demonstrates that the agent learned to find the target and then navigate toward the target while keeping the target centered in the field of view. }
\label{fig:ppo_tracker_frames}
\end{figure}

Figures \ref{fig:ppo_sreacher_frames}, \ref{fig:ppo_reacher_frames} and \ref{fig:ppo_tracker_frames} show 3 episodes of evaluation rollouts for PPO agents at the end of training on the static reacher, reacher and tracker environments. The roll-outs are depicted with two images captured on 5-second intervals. The embedded image is the observation received by the agent and the larger image is from a separate camera used to record the robot and the target. For each environment, each episode commences with the agent looking away from the target and the monitor. Within the first 5 seconds, the agent locates the target, and once located the agent begins to move towards the target whilst keeping the target as centered as possible in its field of view. This is easiest in the static reacher environment where the target is always in the same location, and most challenging in the reacher environment. 

In Figure \ref{fig:ppo_sreacher_frames}, the pattern of the size and location of the target is very similar in each evaluation roll-out. This suggests that the PPO agent has learned a stable and robust policy. An additional sign of the agent's efficacy on this task is that after just 20 seconds the target size is quite larger. Alternatively, in the reacher environment after 35 seconds the target still appears relatively small and sometimes poorly centered. More learning time is clearly needed to deal with the larger problem. 

Figure \ref{fig:ppo_tracker_frames} highlights the fact that the complexity of the tracker environment is moderate compared to the static reacher and reacher. In particular, the agent consistently locates and centres the target within the first 5 seconds, similar to static reacher and unlike reacher, which sometimes takes more than 5 seconds (depending on the target location.) Although the target is typically centered after 35 seconds, it is not nearly as large as in the case of static reacher. This is attributed to the fact that the agent must learn to approach and centre the drifting target. The precision of the agent is also limited by size of the discrete movements in the environment. Once again, additional training time would allow for further policy refinements on this task.

\subsection{Limitations and Challenges}

The implementation of the RL environment has some limitations, such as the synchronization of the agent and the environment. Moreover, from the agent's perspective time advances in discrete steps. This simplifies the tracker tasks because the target only moves as fast as the agent can select actions. In subsequent versions of the environment, the target will drift at a rate that is independent of the agent's action selection rate. As described above, for example, the agent faces a discrete action space that limits the movement of the robot at each time step to a single joint and a fixed step size. Subsequent versions of the environment will add the option to use a continuous action space. A continuous action space will provide numerous advantages, including the ability to learn a smooth control policy and a wider choice of additional RL algorithms. Finally, the observation space is composed of one image captured by the camera on the end-effector of the robotic arm. A single image state space, however, may not include all of the information relevant to reacher and tracker tasks. In the case of tracker, for example, a single image does not provide information about which direction the target is moving. Thus, the problem can be considered partially observable \cite{krishnamurthy2016partially}. This can be handled by using the combined recent history of state (or states and actions) as the input to the agent's policy \cite{mnih2013playing}, or deep partially observable RL algorithms \cite{igl2018deep}. In addition, the physical robot has limits on its joint positions. Based on image states alone, however, the agent may not have sufficient information to learn these limits. The subsequent implementation of the environment will include the option to include both the current image and joint position in the state space.

Reward calculation from raw images is an open and challenging problem in RL. It will become relevant as more RL is trained in the real-world. In this work, the reward is defined based on the size and centering of the target in the frame. Unlike the previous work, the background in the image is complex, the lighting is inconsistent and the agent has the ability to look away from the monitor containing the target. As a result, image processing needs to be applied at each time step to determine if a reward should be given. This represents additional computational resources and requires problem-specific engineering of the reward function. Moreover, due to the limitations of image processing under these conditions, the rewards issued to the agent are noisy thereby increasing the learning challenge. 

DRL algorithms suffer from poor sample efficiency \cite{yu2018towards}. Sample inefficiency occurs in two contexts: \textit{a}) train RL policy from scratch requires the agent to collect a large number of experience tuples (state, action, reward, next state), and \textit{b}) the agent must be trained from scratch multiple times in order to find the best hyper-parameters for the given task. During online training, the robot cannot be used for other productive tasks.  Future work will explore strategies for better robot utilization and sample efficiency.

Sample efficiency is improved with off-policy algorithms that allow for the reuse of experience \cite{van2016deep}, the prioritization of particularly rewarding experience\cite{schaul2015prioritized}, and the distribution of agent learning across multiple robots\cite{gu2017deep}. In this work, however, the sensitivity of DQN to hyper-parameters necessitated significantly more training time than PPO. More generally, sim-2-real offers the potential to initialize the RL via faster, lower-cost interactions with a simulated robot, and then refine the policy on the real-world robot \cite{tan2018sim,yu2020learning}. Similarly, model-based RL can be used for on- and off-policy RL to improve sample efficiency \cite{nagabandi2018learning, yang2020data}.

Other improvements may stem from self-supervised learning to reduce the agent's reliance on heavily engineered, problem-specific rewards \cite{levine2018learning,pinto2016supersizing,ibarz2021train} and the use of frame skipping and observation skipping to reduce observation and decision costs in predictable regions of the state space \cite{lakshminarayanan2017dynamic,bellinger2023dynamic}.

\section{Conclusion}

This work explored the potential of reinforcement learning algorithms to learn optimal control policies for vision-based reacher and tracker tasks with a UR10e robotic arm. We describe the experimental setting and results of online policy learning with DQN and PPO. Whereas much of the research on vision-based RL is undertaken in highly controlled settings, here we train the agents in a complex physical setting with variable light conditions. We find that in these conditions the image-based reward calculations can be very sensitive to the background and variable light leading to a noisy reward signal. Careful image processing prior to reward calculation is required to minimize the risk of noise that can harm policy learning. 

Our results show that proximal policy optimization learns a better, more stable policy with less data than deep Q-learning. In order to improve its performance, we expect that DQN requires a more sophisticated exploration strategy and non-uniform sampling of its replay buffer to focus updates on the most informative experience. Finally, we highlight general challenges and future directions for the study of reinforcement learning in vision-based open-world robotics as: i) reducing the reliance on heavily engineered, domain-dependent rewards, ii) lowering the dependence on computationally expensive image processing for reward calculation, and iii) improving the robustness to the potential for degraded learning due to a noisy reward signal resulting from imprecision in the imaged-based reward calculations.

\bibliographystyle{alpha}
\bibliography{library}

\end{document}